\documentclass[conference]{llncs} 

\usepackage{amsfonts}
\usepackage{amsmath} 
\usepackage{amssymb}  
\usepackage{graphicx}
\usepackage{graphics} 
\usepackage{mathptmx} 
\usepackage{url}
\usepackage{rotating}
\usepackage{subfigure}
\usepackage{wrapfig}
\usepackage{multicol}
\usepackage[linesnumbered,boxruled,lined]{algorithm2e}
\newtheorem{example2}{\bf Example}
\newcommand{\rif}{\stackrel{\,\,+}{\leftarrow}}

\newenvironment{s_itemize}{\begin{list}{$\bullet$}
{\setlength{\rightmargin}{0em}
\setlength{\itemsep}{0em}
\setlength{\topsep}{0em}
\setlength{\parsep}{0em}}}{\end{list}}

\newcounter{ctr}

\begin{document}
\title{Mixed Logical and Probabilistic Reasoning for Planning and
  Explanation Generation in Robotics}

\author{Zenon Colaco \and Mohan Sridharan}
                                                                                
\institute{Department of Electrical and Computer Engineering, The
  University of Auckland, NZ\\
  \texttt{zncolaco@gmail.com; m.sridharan@auckland.ac.nz} }

\maketitle

\begin{abstract}
  Robots assisting humans in complex domains have to represent
  knowledge and reason at both the sensorimotor level and the social
  level. The architecture described in this paper couples the
  non-monotonic logical reasoning capabilities of a declarative
  language with probabilistic belief revision, enabling robots to
  represent and reason with qualitative and quantitative descriptions
  of knowledge and degrees of belief. Specifically, incomplete domain
  knowledge, including information that holds in all but a few
  exceptional situations, is represented as a Answer Set Prolog (ASP)
  program. The answer set obtained by solving this program is used for
  inference, planning, and for jointly explaining (a) unexpected
  action outcomes due to exogenous actions and (b) partial scene
  descriptions extracted from sensor input. For any given task, each
  action in the plan contained in the answer set is executed
  probabilistically. The subset of the domain relevant to the action
  is identified automatically, and observations extracted from sensor
  inputs perform incremental Bayesian updates to a belief distribution
  defined over this domain subset, with highly probable beliefs being
  committed to the ASP program. The architecture's capabilities are
  illustrated in simulation and on a mobile robot in the context of a
  robot waiter operating in the dining room of a restaurant.
\end{abstract}

\section{Introduction}
\vspace{-0.75em}
\label{sec:intro}
Robots collaborating with humans in complex domains receive far more
raw sensor data than can be processed in real-time.  The information
extracted from the sensor inputs can be represented probabilistically
to quantitatively model the associated uncertainty (``$90\%$ certain I
saw the book on the shelf'').  Robots also receive useful commonsense
knowledge that is difficult to represent quantitatively (``books are
usually in the library''), and human participants may not have the
time and expertise to provide elaborate and accurate feedback. To
collaborate with humans, these robots thus need to represent knowledge
and reason at both the cognitive level and the sensorimotor level.
This objective maps to fundamental research challenges in knowledge
representation and reasoning. The architecture described in this paper
exploits the complementary strengths of non-monotonic logical
reasoning and probabilistic belief revision as a significant step
towards addressing these challenges.  Specifically, the commonsense
logical reasoning capabilities of Answer Set Prolog (ASP), a
declarative language, is coupled with probabilistic belief updates, to
support the following key features:
\begin{s_itemize}
\item An ASP program represents incomplete domain knowledge, including
  information that holds in all but a few exceptional situations. The
  \emph{answer set} obtained by solving the ASP program is used for
  planning and jointly (a) explaining unexpected action outcomes by
  reasoning about exogenous actions; and (b) identifying object
  occurrences that best explain partial scene descriptions obtained
  from sensor inputs.
\item For any given task, each action in the plan created by inference
  in the ASP program is executed probabilistically. The relevant
  subset of the domain (for this action) is identified automatically,
  and the sensor observations perform incremental Bayesian updates to
  a belief distribution defined over this subset of the domain,
  committing highly probability beliefs as statements to the ASP
  program.
\end{s_itemize}
The architecture thus enables robots to represent and reason with
qualitative and quantitative descriptions of knowledge and degrees of
belief. In this paper, the architecture's capabilities are
demonstrated in simulation and on a mobile robot, in the context of a
robot waiter operating in the dining room of a restaurant.

\vspace*{-0.75em}
\section{Related Work}
\vspace{-0.75em}
\label{sec:relwork}
Knowledge representation, planning and explanation generation are
well-researched areas in robotics and artificial intelligence.
Logic-based representations and probabilistic graphical models have
been used to plan sensing, navigation and interaction for robots and
agents. Formulations based on probabilistic representations (by
themselves) make it difficult to perform commonsense reasoning, while
classical planning algorithms and logic programming tend to require
considerable prior knowledge of the domain and the agent's
capabilities, and make it difficult to merge new, unreliable
information with an existing knowledge base. For instance, the
non-monotonic logical reasoning capabilities of
ASP~\cite{gelfond:aibook14} have been used for tasks such as reasoning
by simulated robot housekeepers~\cite{erdem:ISR12} and coordination of
robot teams~\cite{saribatur2:iros14}. However, ASP does not support
probabilistic analysis of uncertainty, whereas a lot of information
extracted from sensors and actuators on robots is represented
probabilistically.

Approaches for generating explanations (e.g., through abductive
inference or plan diagnosis) use the systems description and
observations of system behavior to explain unexpected
symptoms~\cite{gelfond:aibook14,reiter:AIJ87}, or use weaker system
descriptions and depend on heuristic representation of intuition and
past experience~\cite{meadows:aaaiws13,ng:pkrr92}.  Probabilistic and
first-order logic-based representations have been combined for better
abductive inference~\cite{raghavan:ecml11}.  Researchers have also
designed architectures for robots that combine deterministic and
probabilistic algorithms for task and motion
planning~\cite{kaelbling:IJRR13}, combine declarative programming and
continuous-time planners for path planning in robot
teams~\cite{saribatur2:iros14}, or combine a probabilistic extension
of ASP with partially observable Markov decision processes (POMDPs)
for human-robot dialog~\cite{zhang:aaai15}.  Some principled
algorithms that combine logical and probabilistic reasoning include
Markov logic network~\cite{richardson:ML06}, Bayesian
logic~\cite{milch:bookchap07}, and probabilistic extensions to
ASP~\cite{baral:TPLP09,lee:aaaisss15}. However, algorithms based on
first-order logic do not provide the desired expressiveness for
modeling uncertainty, e.g., it is not always possible to express
degrees of belief quantitatively. Algorithms based on logic
programming do not support one or more of the desired capabilities
such as incremental revision of (probabilistic) information; reasoning
as in causal Bayesian networks; and reasoning with large probabilistic
components. Towards addressing these challenges, our prior work
developed architectures that couple declarative programming and
probabilistic graphical models for logical inference, deterministic
and probabilistic planning on robots~\cite{zhang:icsr14,zhang:TRO15}.
This paper retains the coupling between logical and probabilistic
reasoning but significantly expands the capabilities of these
architectures to support: (1) explanation of unexpected action
outcomes and the partial descriptions extracted from sensor inputs;
and (2) representation and reasoning at a higher resolution using ASP,
making the probabilistic reasoning more computationally efficient.

\vspace{-0.5em}
\section{Proposed Architecture}
\begin{figure}[tbc]
  \begin{center}
    \includegraphics[width=0.65\columnwidth]{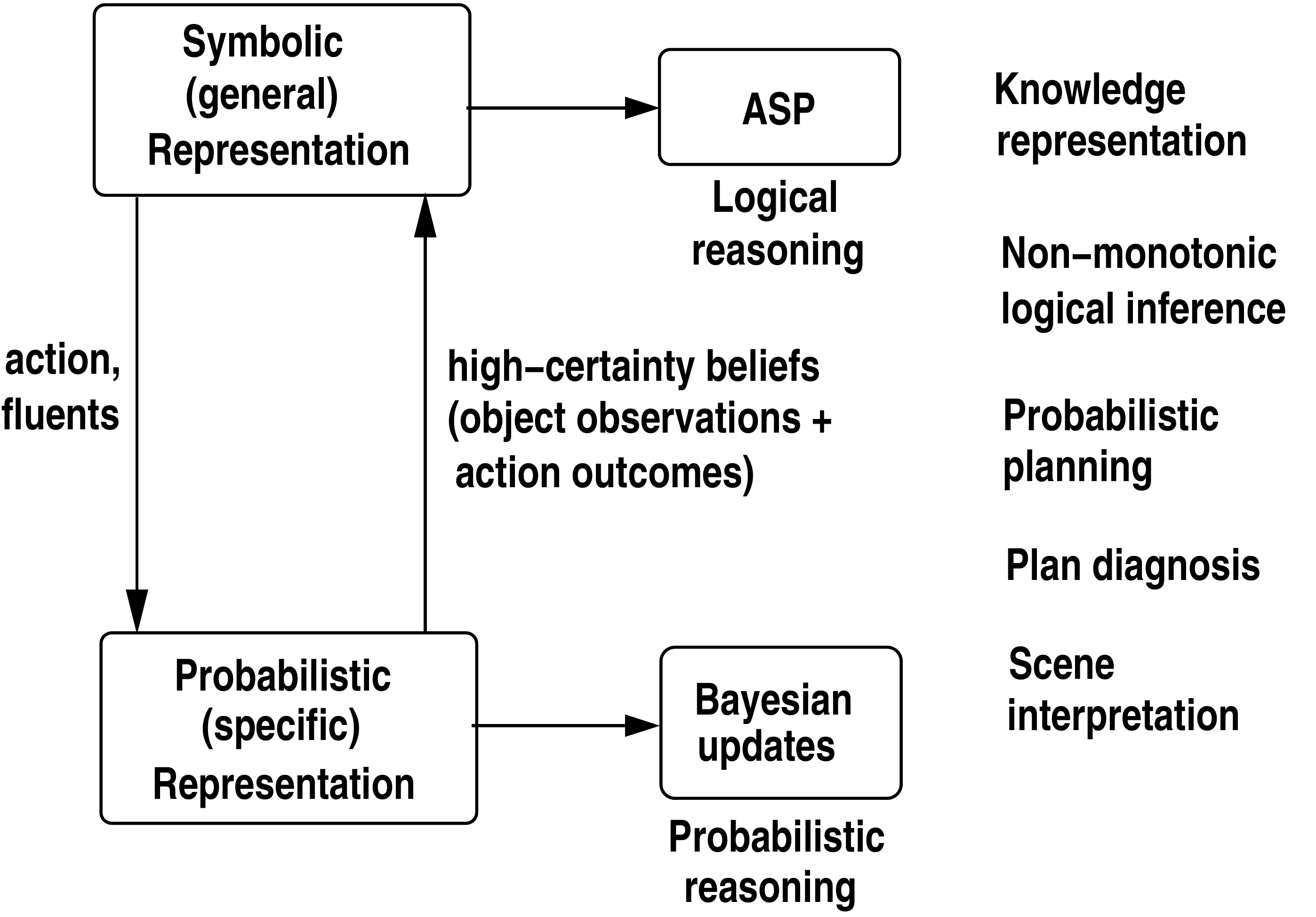}
  \end{center}
  \vspace{-2em}
  \caption{An overview of the architecture that combines the
    complementary strengths of declarative programming and
    probabilistic graphical models for inference, planning, and
    diagnosis.}
  \label{fig:arch-overview}
  \vspace{-1.5em}
\end{figure}

\vspace{-0.75em}
\label{sec:arch}
Figure~\ref{fig:arch-overview} is an overview of the mixed
architecture. The symbolic representation is translated to an Answer
Set Prolog (ASP) program used for non-monotonic logical inference and
planning a sequence of actions for any given task. For each action,
the relevant subset of the domain is defined automatically. Sensor
observations perform incremental Bayesian updates to a probability
distribution over this domain subset, committing high probability
beliefs (action outcomes, observation of object attributes) as
statements to the ASP program.  Observed unexpected action outcomes
are explained by reasoning about exogenous actions, and objects are
identified to best explain the partial descriptions extracted from
visual cues.  ASP-based representation and reasoning is performed at a
resolution that provides high reliability while also simplifying the
(coupled) probabilistic reasoning and tailoring it to specific
actions.

The syntax, semantics and representation of the transition diagrams of
the architecture's domain representation are described in an
\emph{action language} AL~\cite{gelfond:aibook14}.  AL has a sorted
signature containing three \emph{sorts}: $statics$, $fluents$ and
$actions$.  Statics are domain properties whose truth values cannot be
changed by actions, fluents are properties whose values are changed by
actions, and actions are elementary actions that can be executed in
parallel.  AL allows three types of statements: \vspace{-0.5em}
\begin{align*}
  &a~~\mathbf{causes}~~l_{b}~~\mathbf{if}~p_0,\ldots,p_m\qquad \qquad
  \qquad\qquad~~~~\textrm{(Causal law)} \\ \nonumber 
  &l~~\mathbf{if}~~p_0,\ldots,p_m\qquad \qquad \qquad \qquad\qquad\qquad~~~~~\textrm{(State constraint)}\\\nonumber 
  &\mathbf{impossible}~~a_0,\ldots,a_k~~\mathbf{if}~~p_0,\ldots,p_m\qquad\qquad~~~\textrm{(Executability condition)}
\vspace{-2em}
\end{align*}
where $a$ is an action, $l$ is a literal, $l_{b}$ is a basic fluent
(also called inertial fluent) literal, and $p_0,\ldots,p_m$ are domain
literals (any domain property or its negation). A collection of
statements of AL forms a system description.

As an illustrative example used in this paper, consider a robot waiter
that greets and seats people at tables in a restaurant, and delivers
orders.  The sorts of the domain are arranged hierarchically, e.g.,
$location$ and $thing$ are subsorts of $entity$; $animate$ and
$inanimate$ are subsorts of $thing$; $person$ and $robot$ are subsorts
of $animate$; $object$ is a subsort of $inanimate$; and $room$,
$area$, $door$, and $floor$ are subsorts of $location$. We include
specific rooms, e.g., $kitchen$ and $dining$, and consider objects of
sorts such as $table$, $chair$ and $plate$, to be characterized by
attributes $size$, $color$, $shape$, and $location$. The sort $step$
is included for temporal reasoning.

\vspace*{-0.75em}
\paragraph{\textbf{ASP Domain Representation:}}
The ASP program is based on a domain representation that includes a
system description $\mathcal{D}_H$ and a history with defaults
$\mathcal{H}$.  $\mathcal{D}_H$ has a sorted signature
$\Sigma_H=\langle\mathcal{O}, \mathcal{F}, \mathcal{P}\rangle$ that
defines the names of objects, functions, and predicates available for
use, and axioms that describe a transition diagram $\tau_H$. Fluents
are defined in terms of the sorts of their arguments, e.g.,
$has\_location(thing, location)$ $in\_hand(robot, object)$,
$is\_open(door)$, and $can\_move(robot, location)$.  The first three
are \emph{basic fluents} that obey the laws of inertia and can be
changed directly by actions; the last one is a \emph{defined fluent}
that is not subject to inertia and cannot be changed directly by an
action. Statics such as $connected(location, location)$ and
$belongs(location, location)$ specify connections between locations,
relation $holds(fluent, step)$ implies a specific fluent holds at a
specific timestep, and $occurs(action, step)$ (hypothesizes) that a
specific action occurs at a specific timestep.  We include
\emph{actions}, e.g., $move(robot, location)$, $seat\_person(robot,
person, table)$, $search\_person(robot, area)$, $pickup(robot,
object)$, $putdown(robot, object)$, and define domain dynamics using
causal laws such as: \vspace*{-0.75em}
\begin{align}
  &move(R, L)~~\mathbf{causes}~~has\_location(R, L) \\ \nonumber
  &pickup(R, O)~~\mathbf{causes}~~in\_hand(R, O)\\ \nonumber
  &open(R, D)~~\mathbf{causes}~~is\_open(D)\\ \nonumber
  &seat\_person(R, P, T)~~\mathbf{causes}~~has\_location(P, L)~\mathbf{if}~has\_location(T, L)
\end{align}
state constraints such as:
\vspace{-1em}
\begin{align}
  &has\_location(O, L)~~\mathbf{if}~~has\_location(R, L),~~in\_hand(R, O) \\\nonumber
  &\neg has\_location(Th, L_2)~~\mathbf{if}~~has\_location(Th, L_1),~L_1\neq L_2 \\\nonumber
  &has\_location(Th, L_2)~~\mathbf{if}~~has\_location(Th, L_1),~belongs(L_1, L_2)\\\nonumber
  &can\_move(R, L_2)~~\mathbf{if}~~has\_location(R, L_1),~connected(L_1, L_2)
\end{align}
and executability conditions such as:
\vspace{-0.75em}
\begin{align}
  &\mathbf{impossible}~~move(R, L)~~\mathbf{if}~~has\_location(R, L) \\\nonumber
  &\mathbf{impossible}~~pickup(R, O)~~\mathbf{if}~~has\_location(R, L_1),~has\_location(O, L_2), L_1\neq L_2  \\\nonumber
  &\mathbf{impossible}~~open(R, D)~~\mathbf{if}~~is\_open(D)
\end{align}
Since robots frequently receive \emph{default} domain knowledge that
is true in all but a few exceptional situations, the domain history
$\mathcal{H}$, in addition to $hpd(action,step)$ and
$obs(fluent,boolean,step)$, the occurrence of specific actions and the
observation of specific fluents at specific time steps, contains
prioritized defaults describing the values of fluents in their initial
states. For instance, it may be initially believed that dishes to be
delivered are typically on a table between the kitchen and the dining
room---if they are not there, they are still in the kitchen. Existing
definitions of entailment and consistency are used to reason with such
histories, and any observed exceptions~\cite{zhang:icsr14}.

The domain representation is translated into a program
$\Pi(\mathcal{D}_H,\mathcal{H})$ in CR-Prolog that incorporates
consistency restoring rules in ASP~\cite{gelfond:aibook14}. $\Pi$
includes the causal laws of $\mathcal{D}_H$, inertia axioms, closed
world assumption for actions and defined fluents, reality checks, and
records of observations, actions and defaults from $\mathcal{H}$.
Every default is turned into an ASP rule and a consistency-restoring
(CR) rule that allows us to assume the default's conclusion is false
to restore $\Pi$'s consistency. ASP is based on stable model
semantics, introduces concepts such as default negation and epistemic
disjunction, and represents recursive definitions, defaults, causal
relations, and language constructs that are difficult to express in
classical logic formalisms. The ground literals in an \emph{answer
  set} obtained by solving $\Pi$ represent beliefs of an agent
associated with $\Pi$---statements that hold in all such answer sets
are program consequences. Inference and planning can be reduced to
computing answer sets of program $\Pi$ by adding a goal, a constraint
stating that the goal must be achieved, and a rule generating possible
future actions.

Our architecture supports reasoning about exogenous actions to explain
the unexpected (observed) outcomes of actions. For instance, to reason
about a door between the kitchen and the dining room being locked by a
human, and to reason about a person moving away from a known location,
we introduce exogenous actions $locked(door)$ and $moved\_from(person,
location)$ respectively, and suitably add (or revise) axioms:
\vspace{-0.75em}
\begin{align}
  \label{eqn:expl-ab}
  &is\_open(D)~~\leftarrow~~open(R, D),~\neg ab(D)\\\nonumber
  &ab(D)~~\leftarrow~~locked(D)\\\nonumber
  &\neg has\_location(P, L)~~\leftarrow~~moved\_from(P, L),~has\_location(P, L)
\end{align}
where a door is considered \emph{abnormal} if it has been locked, say
by a human. We also introduce an \emph{explanation generation} rule
and a new relation $expl$:
\vspace{-0.75em}
\begin{align}
  \label{eqn:expl}
  &occurs(A, I)~|~\neg~occurs(A, I)~\leftarrow~exogenous\_action(A),~I~<~n \\\nonumber
  &expl(A, I)~\leftarrow~action(exogenous, A),~occurs(A, I),~not~hpd(A, I)
\end{align}
where $expl$ holds if an exogenous action is hypothesized but there is
no matching record in the history. We also include \emph{awareness}
axioms and \emph{reality check} axioms: \vspace{-0.75em}
\begin{align}
  &holds(F, 0)~~or~~\neg~holds(F, 0)~\leftarrow~fluent(basic, F)~~~~~~\%~awareness~axiom\\ \nonumber
  &occurs(A, I)~\leftarrow~hpd(A, I)\\ \nonumber
  &\leftarrow~~obs(fluent, true, I),~\neg~holds(fluent, I)~~~~~~~~~~~~\quad\%~reality~check\\ \nonumber
  &\leftarrow~~obs(fluent, false, I),~holds(fluent, I)~~~~~~~~~~~~~~\quad\%~reality~check
\end{align}
The reality check axioms cause a contradiction when observations do
not match expectations, and the explanation for such unexpected
symptoms can be extracted from the answer set of the corresponding
program~\cite{gelfond:aibook14}. The new knowledge is included and
used to generate the subsequent plans. This approach provides
\emph{all} explanations of an unexpected symptom---using a CR rule
instead of the explanation generation rule (in
Statement~\ref{eqn:expl}) provides the minimal explanation (see
below).

A robot processing sensor inputs (e.g., camera images) is typically
able to extract partial descriptions of scene objects. The proposed
architecture also identifies object occurrences that best explain
these partial descriptions. In our illustrative example, we introduce
static relations to establish object class
membership\footnote{Relation $member(object, class)$ is applied
  recursively in a class hierarchy, $is\_a(object, class)$ denotes an
  instance of a specific class, and $class\_known(object)$ holds for
  any object whose class label is known.} and introduce relations to
capture ideal (and default) definitions of object attributes, e.g.,
for a table: \vspace{-0.75em}
\begin{align}
  \label{eqn:chair}
  &has\_color(O, white)~\leftarrow~member(O, table)\\\nonumber
  &has\_size(O, medium)~\leftarrow~member(O, table)\\\nonumber
  &has\_wheels(4)~\leftarrow~member(O, table),~\neg~has\_location(O,
  kitchen)
\end{align}
where tables usually have wheels except in the kitchen. Similarly, for
a chair:
\vspace{-0.75em}
\begin{align}
  \label{eqn:desk}
  &has\_color(O, white)~\leftarrow~member(O, chair)\\\nonumber  
  &has\_size(O, medium)~\leftarrow~member(O, chair)\\\nonumber
  &\neg has\_wheels(O)~\leftarrow~member(O, chair)
\end{align}
Other objects and object attributes are encoded similarly.  As before,
a reality check axiom causes an inconsistency when an object does not
have a class label due to incomplete information, and a CR rule
restores consistency by assigning class labels:
\vspace{-0.75em}
\begin{align}
  &\leftarrow~object(O),~not~class\_known(O)~~~~\%~reality~check\\\nonumber
  &is\_a(O, C)~\rif~~object(O)~~~~~~~~~~~~~~\qquad\%~CR~rule
\end{align}
This assignment of a class label to an object is based on the smallest
number of rules that need to be relaxed to support the assignment.
This information is also added to the ASP program and be used for
subsequent reasoning. However, both planning and object recognition
are based on processing sensor inputs and moving to specific
locations---these tasks are accomplished using probabilistic
algorithms, as described below.

\vspace*{-0.75em}
\paragraph{\textbf{Probabilistic Domain Representation:}}
Our previous work created the ASP-based representation at a coarser
resolution (e.g., rooms and places), and \emph{refined} it by adding
suitable actions, fluents and sorts (e.g., cells in rooms) to define a
transition diagram and create its probabilistic version that was
modeled as a POMDP~\cite{zhang:icsr14}. The proposed architecture
models the domain at a higher resolution using ASP. For any given
task, the ASP-based plan consists of primitive actions that can be
executed by the robot. Each such action is executed probabilistically,
with the robot maintaining a probability (belief) distribution over
the relevant subset of the domain that is identified automatically
based on the set of related fluents, e.g., for moving between two
tables, the robot only needs to reason about its own location in a
subset of areas. The belief distribution is revised incrementally by
sensor observations using Bayesian updates. For instance, to update
the belief about the location of a specific dish in the dining room:
\vspace{-1em}
\begin{align}
  \label{eqn:bayes}
  p(E_i | O_i) = \frac{ p(O_i | E_i) p(E_i) }{ ( p(O_i | E_i) p(E_i) +
    p(O_i | \neg E_i) p(\neg E_i) ) }
\end{align}
where $O_i$ is the event the dish was observed in area $i$, $E_i$ is
the event the dish exists in area $i$, making $p(O_i | E_i)$ and
$p(E_i | O_i)$ are the observation likelihood and the posterior
probability of existence of the dish in each area in the dining room.
The initial knowledge (i.e., prior: $p(E_i)$ is based on domain
knowledge or statistics collected in an initial training phase (see
Section~\ref{sec:expres}). A Bayesian state estimation approach is
used by the robot to estimate its own position (e.g., \emph{particle
  filters}), navigate, and to process sensor inputs to extract
information about objects being observed.

In summary, for any given task, ASP planning provides a plan with
deterministic effects. The first action in the plan and relevant
information (from ASP inference) identify the subset of the domain to
be represented probabilistically, and set the initial (probabilistic)
belief distribution. The action is executed probabilistically,
updating the belief distribution until a high belief indicates action
completion or a time limit is exceeded, and adding relevant statements
to the ASP program. Any unexpected action outcomes and partial scene
descriptions are explained by reasoning about exogenous actions and
possible objects. Once these explanations restore consistency, either
the next action (in the plan) is selected for execution or a new plan
is created.  In what follows, we refer to the proposed architecture as
the ``mixed architecture'', and compare it with two algorithms: (1)
ASP-based reasoning for completing the assigned tasks; and (2) a
(greedy) probabilistic approach that maintains a probabilistic belief
distribution and heuristically selects actions (and makes decisions)
based on the most likely state.

The mixed architecture raises some subtle issues. First, committing
probabilistic beliefs above a specific threshold (e.g., $0.85$) as
fully certain statements to the ASP program may introduce errors, but
the non-monotonic reasoning capability of ASP helps the robot recover.
Second, with previous work that reasoned at a coarser resolution with
ASP and used POMDPs for probabilistic planning (a) computing POMDP
policies for each ASP action is computationally expensive; and (b)
there may be improper reuse of information if the probabilistic belief
distribution is not reset between trials~\cite{zhang:TRO15}. Third,
the mixed architecture presents an interesting trade-off between the
resolution of symbolic representation and probabilistic
representation. Moving most of the reasoning to a symbolic
representation can reduce accuracy and also be computationally
expensive---the mixed architecture is a good trade-off of accuracy and
efficiency.

\vspace*{-0.5em}
\section{Experimental Setup and Results}
\vspace*{-0.75em}
\label{sec:expres}
Experiments were conducted in simulation and on a mobile robot in
scenarios that mimic a robot waiter in a restaurant.  The robot's
tasks include finding, greeting and seating people at tables, and
delivering orders appropriately. 

\subsection{Experimental Setup and Hypotheses}
\vspace{-0.5em}
\label{sec:expres-setup}
The experimental trials used existing implementations of relevant
control and sensor input processing algorithms. In an initial training
phase, the robot collected statistics of executing these algorithms to
compute the motion error models and observation likelihood models.
These models were also used to make the simulation trials more
realistic.

The experimental trials considered three hypotheses, evaluating that
the mixed architecture: (H1) generates plans for different tasks, and
explains unexpected outcomes and partial descriptions extracted from
sensor inputs; (H2) significantly improves the task completion
accuracy and provides similar task completion time in comparison with
just ASP-based reasoning; and (H3) significantly improves task
completion time and provides similar accuracy in comparison with the
purely probabilistic approach. We provide both qualitative and
quantitative results in simulation and on a mobile robot.

\vspace{-0.5em}
\subsection{Experimental Results}
\vspace{-0.5em}
\label{sec:expres-results}
The following execution traces demonstrate the planning and diagnosis
capabilities.
\vspace{-0.25em}
\begin{example2}\label{ex:plan-expl}[Explain unexpected action outcome]\\
  {\rm The task is to return dish $ds_1$ to the kitchen from the
    dining room, e.g., from $table1$ to area $a_3$ in
    Figure~\ref{fig:domain-map}.  Unknown to the robot, the door $d_2$
    has been locked.
    \begin{s_itemize}
    \item The robot is in the dining room with the dish in hand:\\
      $holds(has\_location(robot, dining), 0),
      ~hold(has\_location(robot, area1),0),$\\
      $holds(in\_hand(robot, ds_1), 0)$\\
      The initial plan obtained by computing the answer set is:
      \vspace{-1em}
      \begin{align*}
        &occurs(move(robot, a_2), 1),~~occurs(move(robot, d_2), 2),\\\nonumber
        &occurs(open(robot, d_2), 3),~~occurs(move(robot, a_3), 4),\\\nonumber
        &occurs(putdown(robot, ds_1), 5).
        \vspace{-1.5em}
      \end{align*}
      
    \item Each step in this plan, starting with the first one, is
      executed probabilistically.

    \item Unfortunately, the robot's attempt to open door $d_2$ does
      not produce the expected observation---instead,
      $obs(is\_open(d_2), false, 3)$ is added to the history.

    \item Diagnosis provides the explanation $expl(locked(d_2), 3)$,
      which invokes Statement~\ref{eqn:expl-ab} to restore
      consistency.

    \item The robot seeks human help to unlock the door, before
      creating a new plan to successfully return the dish to the
      kitchen.
    \end{s_itemize}
  }
\end{example2}
\vspace{-1em}
\begin{example2}\label{ex:scene-expl}[Explain partial scene description]\\
  {\rm The robot delivering a dish sees a medium-sized white object from
    a distance, but is unable to assign a class label in the absence
    of any further information.
  \begin{s_itemize}
  \item Then initial knowledge consists of:\\
    $has\_size(ob_1, medium),~has\_color(ob_1, white)$

  \item Two possible interpretations are generated by
    Statements~\ref{eqn:chair},\ref{eqn:desk} as an explanation:\\
    $is\_a(ob_1, table)$ or $is\_a(ob_1, chair)$.

  \item As the robot gets closer, it observes: $has\_wheel(ob_1,
    4),~has\_location(ob_1, dining)$, i.e., the object has wheels and
    is in the dining room.  Statements~\ref{eqn:chair},\ref{eqn:desk}
    provide the interpretation: $is\_a(ob_1, table)$.
  \end{s_itemize}
}
\end{example2}

\vspace{-1em}
\paragraph{\textbf{Simulation Experiments:}}
The robot in the simulator had to find and move (seat) dishes (people)
to specific locations.  Table~\ref{tab:simulation_exp} summarizes the
results---each entry is the average of $500$ trials. The task, initial
position of the robot, and position of objects, were different between
the trials, and paired trials were used to establish statistical
significance. In each paired trial, for each approach being compared,
the initial location of the robot and the location of domain objects
are the same. The statistically significant results show that the
mixed architecture (a) significantly improves the accuracy and
provides similar task completion time in comparison with ASP-based
reasoning; and (b) significantly improves task completion time and
provides similar accuracy in comparison with the probabilistic
approach. Furthermore, the planning and execution time are
significantly reduced in comparison with approaches that combined ASP
with probabilistic graphical models~\cite{zhang:icsr14}, while
providing comparable accuracy---analysis in other domains may help
automate the choice of resolution for symbolic and probabilistic
representations for a given task and domain.

\begin{table}[tbc]
  \begin{center}
    \caption{Task completion accuracy and time using only ASP, and
      using a probabilistic approach, expressed as a factor of the
      values provided by the mixed architecture. The proposed mixed
      architecture \emph{significantly} reduces the task completion
      time while improving the accuracy.}
  \begin{tabular}{|c|c|c|} \hline
    Algorithms & \multicolumn{2}{|c|}{Evaluation metrics}\\ \cline{2-3}
     & ~~~Accuracy~~~ & ~~~Time~~~ \\ \hline
    ASP only & $0.82$ & $1.06\pm 0.6$ \\ \hline
    Probabilistic  & $0.99$ & $3.32\pm 3.0$ \\ \hline
    Mixed approach & $1$ & $1$ \\ \hline
  \end{tabular}
  \vspace{-1.5em}
  \label{tab:simulation_exp}
  \end{center}
\end{table}

\vspace{-0.75em}
\paragraph{\textbf{Robot Experiments:}}
Trials were conducted on a \emph{Turtlebot} (Figure~\ref{fig:robot})
equipped with a Kinect (RGB-D) sensor, range sensors, and an on-board
processor running Ubuntu Linux. Our architecture and algorithms were
implemented using the Robot Operating System (ROS). Trials included
instances of the domain introduced in Section~\ref{sec:arch}, each
with one or more tables, people and other objects, e.g.,
Figure~\ref{fig:domain-map}. The robot was equipped with probabilistic
algorithms to determine the attribute values of objects (e.g., color
and shape) from camera images, revise the map of the domain, and
determine its location in the map. The robot was able to use the mixed
architecture to successfully complete the assigned tasks in all such
scenarios, with results of paired trials being similar to those
obtained in simulation. A video of an experimental trial showing
planning and diagnosis can be viewed online:
\url{https://vimeo.com/130279856}

\begin{figure*}[tbc]
  \begin{center}
    \subfigure[0.35\textwidth][Example domain.]{
    \includegraphics[width=0.35\columnwidth]{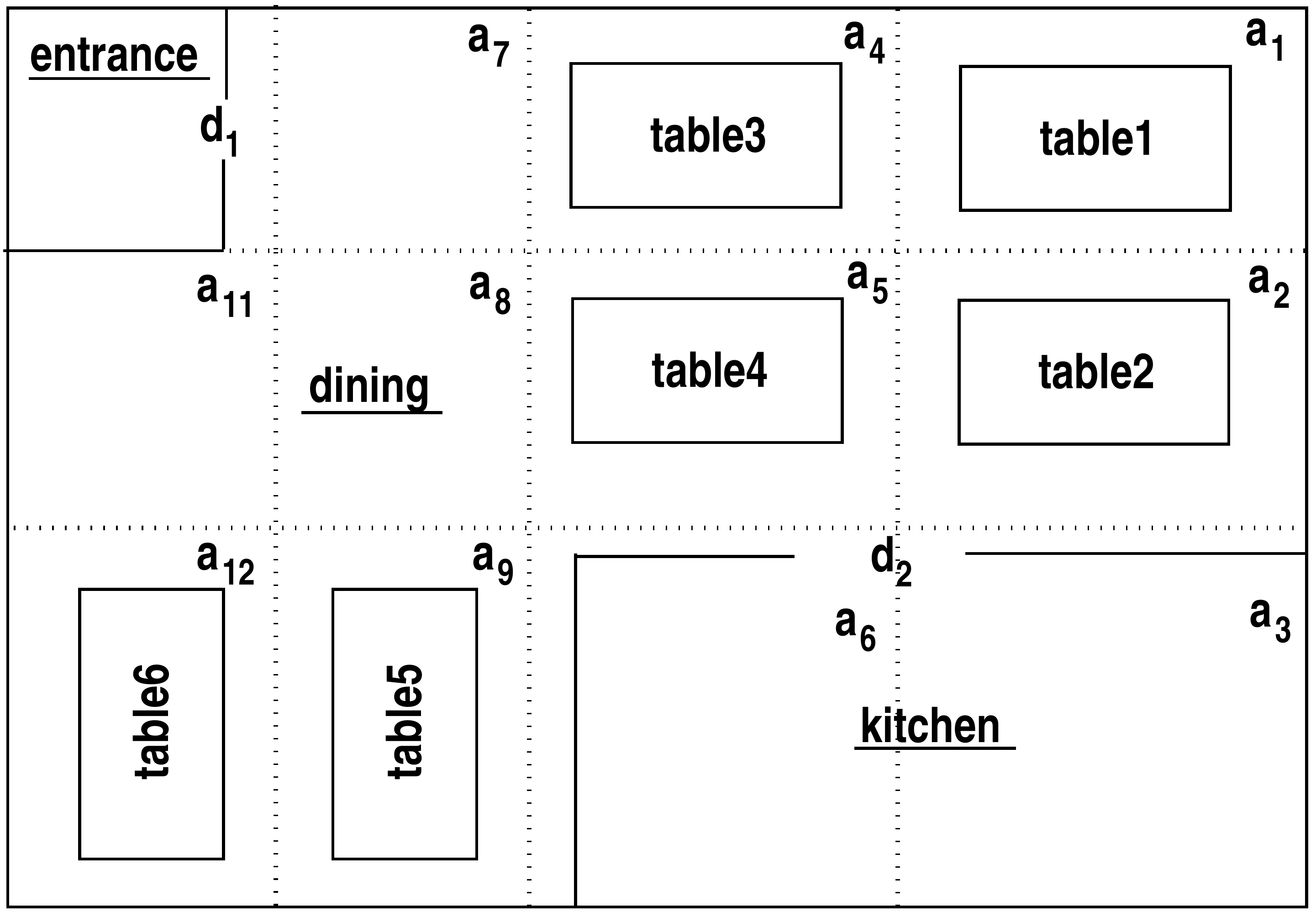}
    \label{fig:domain-map}
    }\hspace{0.2in}
    \subfigure[0.25\textwidth][Turtlebot robot.]{
      \includegraphics[height=0.25\columnwidth]{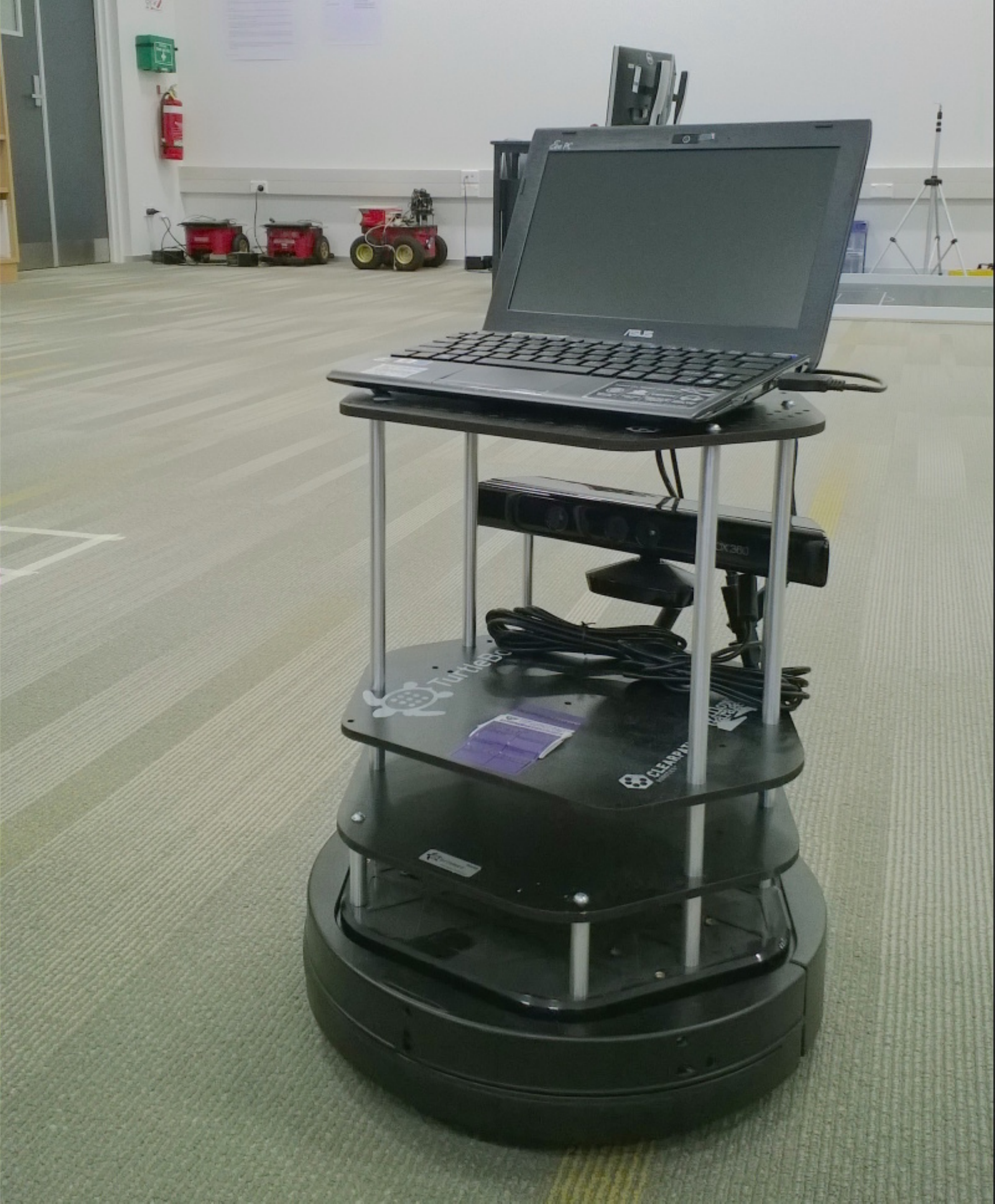}
      \label{fig:robot}
    }
    \vspace{-1em}
    \caption{(a) Example map of illustrative domain used for
      experimental evaluation, with rooms, doors, and tables (people
      and robot not shown); and (b) the Turtlebot mobile robot
      platform.}
    \label{fig:domain-robot}
  \end{center}
  \vspace{-2.5em}
\end{figure*}

\vspace{-0.75em}
\section{Conclusions}
\vspace{-1em}
\label{sec:conclusion}
This paper described an architecture that mixes the complementary
strengths of declarative programming and probabilistic belief updates.
Plans created using ASP-based non-monotonic logical reasoning are
implemented probabilistically, with high probability observations and
action outcomes adding statements to the ASP program.  The
architecture enables a robot to explain unexpected action outcomes by
reasoning about exogenous actions, and to identify objects that best
explain partial scene descriptions.  These capabilities have been
demonstrated through experimental trials in simulation and on a mobile
robot in scenarios that mimic a robot waiter in a restaurant's dining
room.  Future work will further investigate the tight coupling and
transfer of control between the logical and probabilistic
representations, with the long-term objective of enabling robots to
collaborate with humans in complex application domains.

\vspace{-0.5em}
\section*{Acknowledgments}
\vspace{-0.75em} The authors thank Michael Gelfond and Rashmica Gupta
for discussions that contributed to the development of the
architecture described in this paper. This work was supported in part
by the US Office of Naval Research Science of Autonomy award
N00014-13-1-0766. All opinions and conclusions described in this paper
are those of the authors.




\end{document}